\documentclass{article}
\usepackage{graphicx} 
\usepackage{algorithm}
\usepackage{algorithmicx}
\usepackage{algpseudocode}
\usepackage{amsmath}
\usepackage{listings}
\usepackage{xspace}
\usepackage{subcaption}
\usepackage{changepage}
\usepackage{authblk}
\usepackage{float}
\usepackage[margin=1in]{geometry}
\usepackage{todonotes}
\usepackage{booktabs}
\usepackage{threeparttable}
\usepackage{array}
\usepackage{wrapfig}
\usepackage[style=numeric, sorting=none, backend=biber, maxnames=3, giveninits=true]{biblatex}
\DeclareNameAlias{sortname}{family-given}
\DeclareNameAlias{default}{family-given}
\usepackage[hidelinks]{hyperref}
\DeclareUnicodeCharacter{202F}{}
\usepackage{enumitem}
\setlist[description]{style=nextline, leftmargin=0pt, labelindent=0pt, labelsep=.5em}

\newcommand{\surfer}{Surfer~2\xspace}
\newcommand{\orchestrator}{Orchestrator\xspace}
\newcommand{\holo}{Holo1.5\xspace}

\title{\vspace*{-2cm} \huge {\bf{\surfer}} \\ {The Next Generation of Cross-Platform \\ Computer Use Agents}\\
\vspace{1cm}
  \includegraphics[width=0.4\textwidth]{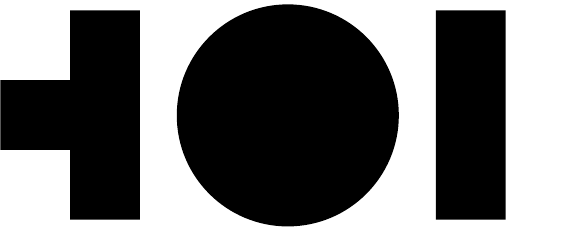}
}

\author{
M.~Andreux,
M.~Bakler,
Y.~Barbier,
H.~Benchekroun,
E.~Biré,
A.~Bonnet,
R.~Bordie,
N.~Bout,
M.~Brunel,
A.~Cambray,
P.-L.~Cedoz,
A.~Chassang,
G.~Cloix,
E.~Connelly,
A.~D.~Constantinou,
R.~De~Coster,
H.~de~La~Jonquière,
A.~Delfosse,
M.~Delpit,
A.~Deprez,
A.~Derupti,
M.~Diaz,
S.~D'Souza,
J.~Dujardin,
A.~Edmund,
M.~Eickenberg,
A.~Fatalot,
W.~Felissi,
I.~Herring,
X.~Koegler,
E.~Le~Jumeau~de~Kergaradec,
A.~Lac,
M.~Langevin,
C.~Lauverjat,
A.~Loison,
A.~Manevich,
A.~Moyal,
A.~Nguyen~Kerbel,
M.~Parovic,
J.~Revelle,
G.~Richard,
M.L.~Richter,
R.~Riochet,
M.~Santos,
R.~Savidan,
L.~Sifre,
M.~Theillard,
M.~Thibault,
I.~Valentini,
T.~Wu,
L.~Yie,
K.~Yuan,
J.~Zubovskij
}

\affil{H Company — Alphabetical order}

\date{October 2025}

\begin{document}

\vspace*{1cm}  

\begin{center}
{\huge \textbf{\surfer} \\ \Large {The Next Generation of Cross-Platform \\ Computer Use Agents}\\
\vspace{1cm}
\includegraphics[width=0.4\textwidth]{images/h-glyph-2025-1.pdf}}
\end{center}

\vspace{1.5em}  

\begin{adjustwidth}{2cm}{2cm}
\begin{center}
{\small
M.~Andreux,
M.~Bakler,
Y.~Barbier,
H.~Benchekroun,
E.~Biré,
A.~Bonnet,
R.~Bordie,
N.~Bout,
M.~Brunel,
A.~Cambray,
P.-L.~Cedoz,
A.~Chassang,
G.~Cloix,
E.~Connelly,
A.~D.~Constantinou,
R.~De~Coster,
H.~de~La~Jonquière,
A.~Delfosse,
M.~Delpit,
A.~Deprez,
A.~Derupti,
M.~Diaz,
S.~D'Souza,
J.~Dujardin,
A.~Edmund,
M.~Eickenberg,
A.~Fatalot,
W.~Felissi,
I.~Herring,
X.~Koegler,
E.~Le~Jumeau~de~Kergaradec,
A.~Lac,
M.~Langevin,
C.~Lauverjat,
A.~Loison,
A.~Manevich,
A.~Moyal,
A.~Nguyen~Kerbel,
M.~Parovic,
J.~Revelle,
G.~Richard,
M.L.~Richter,
R.~Riochet,
M.~Santos,
R.~Savidan,
L.~Sifre,
M.~Theillard,
M.~Thibault,
I.~Valentini,
T.~Wu,
L.~Yie,
K.~Yuan,
J.~Zubovskij
}

\vspace{0.5em}

{\small H Company — Alphabetical order}

\vspace{0.5em}

{\small October 2025}
\end{center}

\vspace{2em}  

\normalsize
\begin{abstract}
Building agents that generalize across web, desktop, and mobile environments remains an open challenge, as prior systems rely on environment-specific interfaces that limit cross-platform deployment. We introduce \surfer, a unified architecture operating purely from visual observations that achieves state-of-the-art performance across all three environments. \surfer integrates hierarchical context management, decoupled planning and execution, and self-verification with adaptive recovery, enabling reliable operation over long task horizons. Our system achieves $97.1$\% accuracy on WebVoyager, $69.6$\% on WebArena, $60.1$\% on OSWorld, and $87.1$\% on AndroidWorld, outperforming all prior systems without task-specific fine-tuning. With multiple attempts, \surfer exceeds human performance on all benchmarks. These results demonstrate that systematic orchestration amplifies foundation model capabilities and enables general-purpose computer control through visual interaction alone, while calling for a next-generation vision language model to achieve Pareto-optimal cost-efficiency.
\end{abstract}
\end{adjustwidth}

\newpage
\begin{figure}[h!]
    \centering
    \makebox[\textwidth][c]{%
        \includegraphics[width=0.97\textwidth]{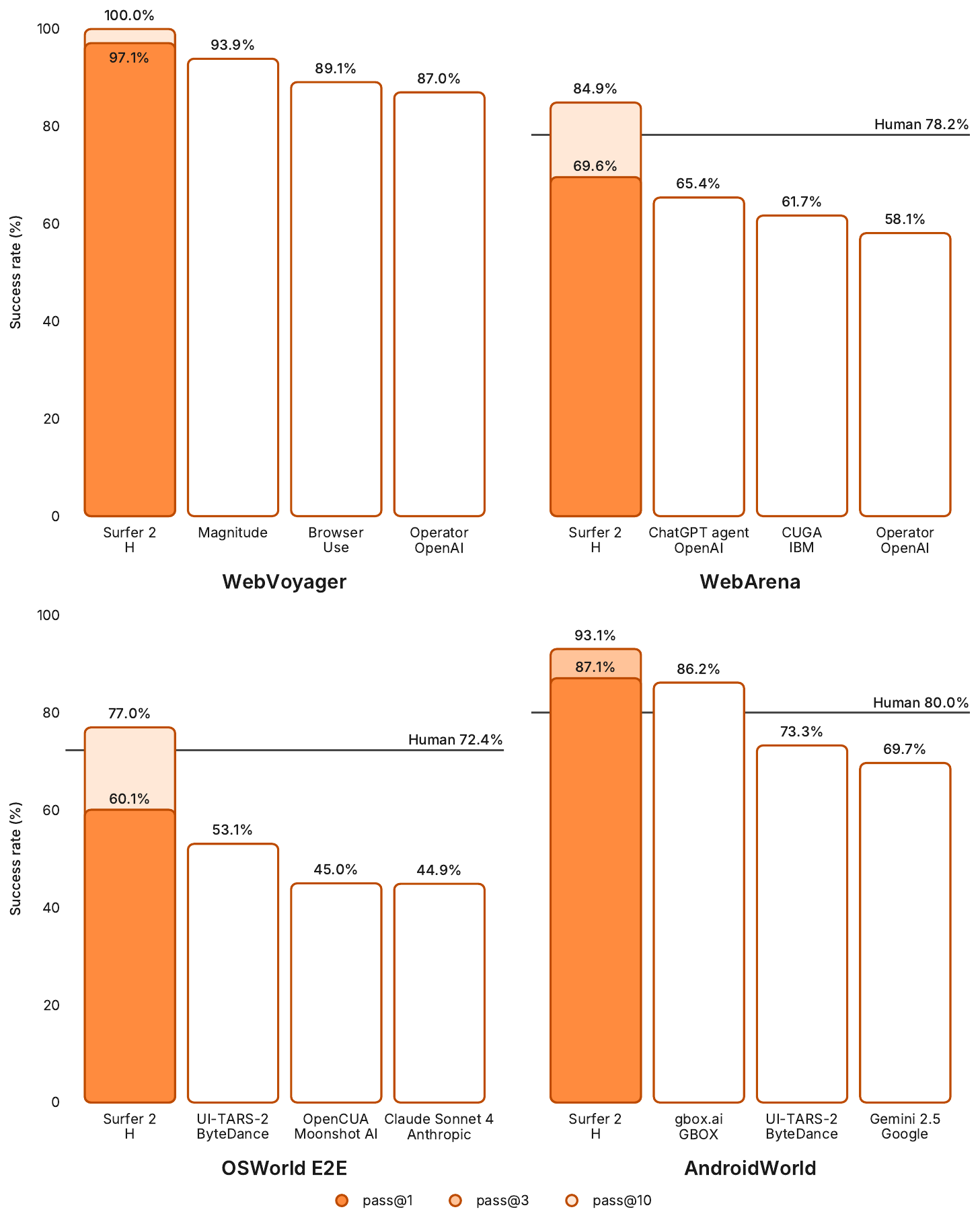}%
    }
    \caption{\surfer state-of-the-art performance on WebVoyager, WebArena, OSWorld E2E, and AndroidWorld. Human performance is indicated when available. }
    \label{fig:benchmarks_square_2}
\end{figure}

\section{Introduction}
Recent advances in Large Language Models (LLMs) and Vision-Language Models (VLMs) unlocked remarkable reasoning capabilities for agentic use cases~\cite{google2025astra}. However, turning these capabilities into reliable, general-purpose agents that operate autonomously in Graphical User Interfaces (GUIs) on complex, real-world tasks remains challenging. In recent years, one dominant path has been to train increasingly large models with minimal scaffolding like tool-call \cite{openai:function-calling:2023, schick2023toolformerlanguagemodelsteach} to improve agentic capabilities \cite{anthropic_sonnet4.5_2025, openai2024gpt4technicalreport, gemini25}. In contrast, this work presents an alternative perspective: with proper orchestration and system design, existing state-of-the-art models can achieve human-level performance and exceed prior systems across multiple benchmarks.

Prior approaches require environment-specific adaptations, such as DOM parsers for web navigation, accessibility trees for mobile interfaces, or specialized APIs for desktop applications, limiting generalization across diverse digital environments. This work introduces \textbf{\surfer}, a unified, hierarchical agent architecture designed for complex tasks across desktop, web, and mobile environments using purely visual interaction.

\surfer comprises three components: \textbf{\orchestrator} (optional high-level planner), a \textbf{Navigator} (low-level GUI executor), and a \textbf{Validator} (evaluation module).
\surfer integrates third-party frontier models and H Company's \holo models~\cite{holo1.5modelcard} in a design that separates long-term strategic planning from short-term tactical execution. A key design insight in \surfer is architectural flexibility. It can enable or bypass the \orchestrator to match task complexity. For long-horizon problems, the \orchestrator runs as a high-level planner in the plan-and-act style \cite{erdogan_plan-and-act_2025}, where it decomposes the user task into verifiable goals, plans ahead, and delegates targeted subtasks to the Navigator. For simple tasks, the \orchestrator is bypassed and the Navigator is invoked directly. Built on our previous SurferH agent \cite{andreux_surfer-h_2025}, the Navigator follows a ReAct (reason+act) loop~\cite{yang_react_2024}. It perceives the environment purely via screenshots, reasons about the next step, and executes constrained keyboard and mouse-level controls with pixel-accurate UI localization from \holo. Upon subtask completion, a Validator inspects the latest screenshots, the execution history, and the proposed answer to assess subtask success in one of two ways: (1) if the \orchestrator is enabled, it leverages the Validator’s report and either advances to the next subgoal or replans accordingly, or (2) if the \orchestrator is disabled, the Validator’s feedback is sent directly to the Navigator, which integrates it into its reasoning and continues the ReAct loop until the task is completed or a termination condition is reached.

Without task-specific fine-tuning, \surfer attains state-of-the-art results on four major benchmarks spanning different Computer Use environments (web browser, desktop, mobile): WebVoyager~\cite{he_webvoyager_2024} and WebArena~\cite{zhou_webarena_2024}, OSWorld~\cite{xie_osworld_2024}, and AndroidWorld~\cite{rawles_androidworld_2025}. On OSWorld and AndroidWorld, \surfer surpasses the human baseline, underscoring that expert agent design is as crucial as model capability.

\section{Related Work}
The development of Computer Use agents capable of controlling computers, web browsers, and mobile devices represents a key frontier in AI. In many real-world scenarios, agents encounter tools and software for which no API or Model Context Protocol (MCP) is available, leaving GUIs as the only viable control surface.  
Consequently, recent research has focused on enabling general-purpose agents to perceive, reason about, and act within GUIs, transforming visual interaction into a universal interface for autonomous computer use.
\subsection{Agents for GUI Control}
The development of agents capable of controlling computers through their graphical interfaces has been a long-standing goal in AI. Early work focused on script-based or rule-based systems for automating specific, repetitive tasks \cite{cypher1993watch}. Subsequent research introduced reinforcement learning (RL) and computer vision techniques for GUI automation \cite{liu2018reinforcement, hsiao2022screenqa}. Recent  advances leverage large language models (LLMs) and vision-language models (VLMs) for more generalized and adaptable control \cite{he_webvoyager_2024, xie_osworld_2024}.

\subsubsection{Browser Use Agents}
Web navigation agents are a highly active research area, with benchmarks like WebVoyager~\cite{he_webvoyager_2024} and WebArena~\cite{zhou_webarena_2024} providing standardized evaluation environments. Early methods often relied on interpreting the Document Object Model (DOM) to understand a page's structure and content~\cite{webgpt, hong_cogagent_2024, zhou_webarena_2024, lavague_docs}. While effective, these text-based approaches struggle with visually-rich elements, dynamic content, or situations where visual layout and context are crucial for task success.

Our work operates on a fundamentally different, multimodal principle: we use image-based states (screenshots) as the primary input, following the approach of Surfer-H~\cite{andreux_surfer-h_2025}. This enables our agent to perceive and interact with the digital environment in a more human-like way, leveraging the visual understanding of large multimodal models (LMMs). Previous works have already explored this path; for instance Set-of-Marks \cite{he_webvoyager_2024} augments screenshots with labeled bounding boxes for each UI element and refers to these labels when issuing clicks. In contrast, our approach operates directly on unaltered screenshots and predicts raw pixel coordinates.
Other approaches apply reinforcement learning to learn skills from scratch \cite{zhou_proposer-agent-evaluatorpae_2024, qi_webrl_2025, yang_react_2024}, whereas we achieve superior performance without task-specific fine-tuning. Our work, similar to \cite{abuelsaad_agent-e_2024}, focuses on architectural improvements rather than model training.

\subsubsection{Desktop Computer Use Agents}
Beyond the web, agents for general Computer Use present unique challenges due to the heterogeneity of application interfaces, multi-app workflows and the requirement for system-level control. OSWorld \cite{xie_osworld_2024} has emerged as a leading benchmark for desktop automation with tasks focusing on Ubuntu, evaluating agents across diverse applications such as LibreOffice, GIMP, VS Code and the OS system. Complementing it, WindowsAgentArena \cite{windows-agent-arena} provides a very similar suite of tasks for Windows-based environments.

Early efforts in Computer Use were open-source, and include OSAtlas \cite{os-atlas} and Aguvis \cite{aguvis}. These established the foundation for developing and evaluating vision-language-action agents capable of operating within general computer environments. Subsequent research, such as \cite{qin_ui-tars_2025, wang_ui-tars-2_2025, ye_mobile-agent-v3_2025, agent-q, opencua, deepminer, pc-agent-e}, has since expanded upon these efforts, exploring diverse architectures and training paradigms to enhance reasoning, perception, and control capabilities. Closest to our work, Agent S3 \cite{gonzalezpumariega2025unreasonableeffectivenessscalingagents} introduces Behavior Best-of-N (bBoN), where it generates multiple parallel trajectories and then selects the most successful one. To make this selection feasible, it first converts dense, raw trajectories into concise ``behavior narratives'' that summarize the agent's actions and their effects. A judge model then compares these narratives to pick the best rollout. Agent S3 can also invoke a coding agent to perform programmatic edits such as bulk operations, file transformations, and structured parsing. 

\subsubsection{Mobile Use Agents}
Over the last two years, the field of mobile agent research has grown with the introduction of the AndroidWorld~\cite{rawles_androidworld_2025} benchmark, which provides a rigorous testbed for agents that require touch-based interactions and multi-app workflows, and the Android in the Wild (AITW) dataset~\cite{rawles_android_2023, li_effects_2024, zhang_you_2024}. Research in this area, including work like \cite{wang_mobile-agent_2024, tang_magicgui_2025, hong_cogagent_2024, bai_digirl_2024, bai_digi-q_2025}, has focused on developing and training specialized models capable of interpreting mobile UIs and executing gestures. While these works demonstrate the power of model-centric approaches, our architecture proves that the same level of performance can be achieved by orchestrating existing models, accommodating the visual and interactive distinctions of mobile platforms.

\subsection{Frontier Models for Agents}
The performance of modern GUI agents is inextricably linked to the capabilities of the underlying frontier models, particularly LLMs and VLMs. Models like GPT-4.1 \cite{openai2024gpt4technicalreport}, o3 \cite{openai_o3_o4mini_system_card_2025}, Claude 4.5 Sonnet \cite{anthropic_sonnet4.5_2025}, and Gemini 2.5 \cite{gemini25} have demonstrated exceptional abilities in multimodal reasoning, zero-shot generalization, and long-context understanding. While many studies have focused on scaling up models \cite{hong_cogagent_2024} or fine-tuning models on large, domain-specific datasets \cite{li_effects_2024, zhang_you_2024}, our work takes a different direction. We employ frontier models and demonstrate that a carefully designed system can achieve state-of-the-art results.

\subsection{Localization Models}
Accurate user interface (UI) element localization remains a key technical challenge for GUI agents operating on visual data. The agent must infer the precise coordinates of a target such as a button, text field, or icon from a screenshot, often conditioned on a natural-language description. This task, known as visual grounding, has motivated the development of specialized vision-language models designed specifically for UI contexts. For instance, UI-TARS~\cite{qin_ui-tars_2025}, \holo~\cite{andreux_surfer-h_2025, holo1.5modelcard}, and CogAgent~\cite{hong_cogagent_2024} are models specifically trained for localizing UI elements. Our system relies on \holo~\cite{holo1.5modelcard}, a specialized localization model, to bridge the gap between the agent's high-level action plan (e.g., ``click the 'Submit' button'') and the pixel-level action required for execution. Our work highlights how effective orchestration and integration of such specialized models are as important as their individual capabilities.

\subsection{Agent Architectures and Learning Paradigms}
Our work on a hierarchical, multi-agent framework builds on established principles in reinforcement learning (RL) and agent design, but it fundamentally differs by employing off-the-shelf models without training. The concept of separating high-level planning from low-level execution has been explored in various contexts \cite{erdogan_plan-and-act_2025, zhou_archer_2024}, including in GUI agent frameworks that use experience-augmented hierarchical planning and internal experience retrieval to address long-horizon tasks \cite{agentS}. A notable example in the mobile domain is the K²-Agent framework \cite{k2_agent_2025}, which explicitly separates a high-level, training-free planner from a low-level, learning-based executor. Our system extends these ideas through a persistent environment state and robust validation mechanisms, ensuring consistency and recoverability across extended workflows. Our use of a planner coordinating sub-agents is a form of multi-agent orchestration, similar in principle to \cite{abuelsaad_agent-e_2024}.

Our system's self-correction and validation loop relates to recent work on autonomous evaluation and refinement of agent behavior \cite{pan_autonomous_2024, yang_react_2024}, particularly in methods that reinforce agents through linguistic feedback and reflective episodic memory rather than weight updates \cite{reflexionLanguageAgents}. The use of an external verification module aligns with research on using language models for critique and dense rewards \cite{cao_beyond_2024}. However, unlike many of these approaches that rely on offline or online RL to update model weights \cite{kumar_conservative_2020, kostrikov_offline_2021, verma_chai_2022, snell_offline_2023, wang_offline_2024, zhai_enhancing_2024, bai_digi-q_2025}, we do not perform any parameter updates. Instead, our findings highlight that coordination and self-correction at the system level can substitute for learning at the model level. We show that even in-context learning, without any gradient updates, can achieve remarkable performance, similar to previous findings~\cite{ma_vision_2024, zhou_memento_2025}.

\section{Agent Architecture}

\begin{figure}[h!]
    \centering
    \includegraphics[width=0.9\textwidth]{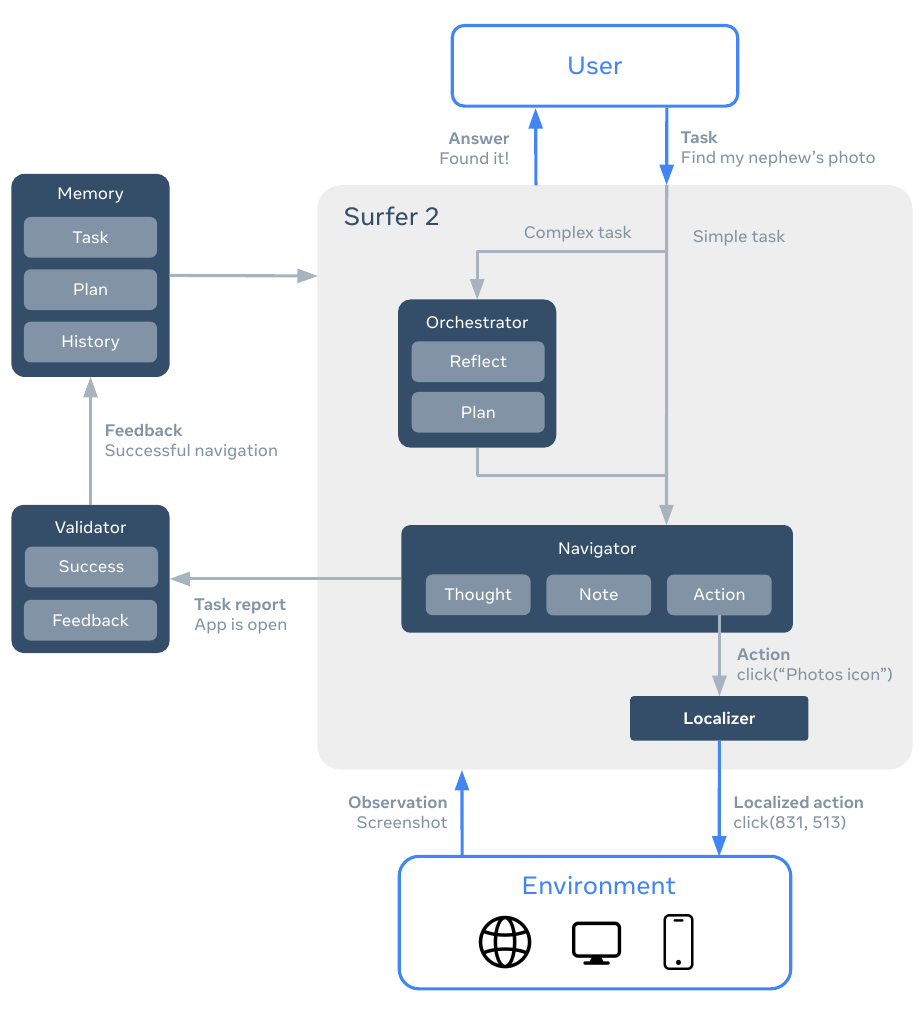}
    \caption{\textbf{\surfer architecture}: an optional \orchestrator plans, the Navigator acts via a \holo Localizer, and a Validator provides feedback.}
    \label{fig:surfer2}
\end{figure}

We outline the components required to build \surfer as shown in Figure~\ref{fig:surfer2}.

\subsection{Design Philosophy}

The design of our architecture follows five core principles.  
First, we adopt a \textbf{separation of concerns}: high-level planning, managed by the \orchestrator, is decoupled from low-level execution handled by the Navigator, allowing each component to specialize and improve independently.  
Second, we employ an automatic \textbf{hierarchical context} mechanism, giving each component access to relevant global information such as the overarching goal, current plan, completed work, and immediate subtask while maintaining scope-specific focus. 
Third, we ensure a \textbf{shared environment state}, in which elements like browser sessions or open applications persist across subtasks and components, enabling incremental progress in dynamic environments.  
Fourth, we emphasize \textbf{explicit validation} through multi-stage verification processes to limit error propagation and promote self-correction.  
Finally, we employ \textbf{chain-of-thought reasoning} with all modules except the Localizer, reasoning explicitly in natural language for better long-horizon performance.

\subsection{\orchestrator}

\paragraph{Main roles.}
The \orchestrator shoulders three key roles simultaneously. As a \textbf{planner}, it decomposes the user goal into sequential, verifiable subtasks and delegates their execution to the Navigator. As a \textbf{coordinator}, it evaluates the Navigator’s outcomes and Validator feedback to detect potential errors or incomplete progress, and replans when necessary to recover from failures. Finally, as a \textbf{communicator}, it decides when to terminate and synthesizes validated results into a coherent final response. The system operates hierarchically: the \orchestrator maintains the global plan and reasoning loop, while the Navigator executes grounded actions within its own local observation–action cycle. This organization ensures efficient task management and robustness through explicit verification and adaptive recovery.

\paragraph{\orchestrator Memory.}
At each decision step, the Orchestrator maintains the overall task objective, the current plan, its execution status, a history of all past interactions with the Navigator, and the latest visual observations from the environment. This persistent state enables the Orchestrator to track progress, learn from previous attempts, and make informed decisions about when to continue, replan, or terminate.

\paragraph{Adaptive planning.}
\surfer employs an adaptive strategy for deciding between single-call execution and multi-step planning based on task complexity heuristics. For simple tasks, the Navigator is directly invoked. For complex tasks involving multiple phases, cross-referencing information, or dependent subtasks, the \orchestrator is triggered to create an explicit plan that breaks the task into a sequence of manageable goals. After each Navigator execution, the \orchestrator analyzes its task report, updates the current goal status, and updates its plan with a mitigation strategy if failures have been detected. The \orchestrator has 4 available actions, which balance high-level planning (\texttt{create\_plan}, \texttt{replan}) with tactical execution (\texttt{delegate}) and external communication (\texttt{answer}).

\subsection{Navigator}
Our Navigator agent is an improved version of the previous agent Surfer-H~\cite{andreux_surfer-h_2025} that now operates across web, desktop, and mobile environments. It extracts relevant information through note-taking, generates action sequences with UI grounding, and verifies task completion through integrated validation. 
Its vision–language policy interprets the current environment state and past trajectory to produce a structured output consisting of a \textbf{note} (information extracted from the latest observation), a \textbf{thought} (reasoning about the next step), and an \textbf{action} (the operation to execute). 
The localizer optionally grounds localizable actions to screen coordinates, enabling interaction with specific UI elements. 
Finally, the environment executes the grounded action and returns a new screenshot observation of the state. 
When the policy issues an \texttt{answer} action to signal task completion, the Validator assesses the result’s completeness and correctness before allowing termination, providing a crucial safeguard against premature or inaccurate responses.

\subsection{Localizer}

The localizer bridges the gap between textual element descriptions (provided by the Navigator) and precise screen coordinates, a critical capability for reliable action execution. It grounds any UI element references in the action to precise screen coordinates, converting textual descriptions like ``blue submit button'' into clickable $(x, y)$ coordinates. This visual grounding problem is fundamental to GUI automation, as even small localization errors can cause actions to miss their intended targets entirely. We use \holo models as Localizer throughout our experiments. 

\subsection{Validator}

Validation is a critical component for preventing premature termination and ensuring answer quality through systematic verification, inspired by prior work on self-reflective agents and feedback-driven verification mechanisms \cite{shinn2023reflexion, agashe2024agents}. 
The Validator examines the Navigator’s complete execution trace including the task specification, reasoning history, sequence of actions, proposed answer, and the most recent $k$ screenshots. It then determines whether the solution satisfies the task requirements based on observable evidence. 
This \textit{VLM-as-a-Judge} operates at two levels of the system hierarchy. 
Within the Navigator, it evaluates each \texttt{answer} action before allowing termination: if validation fails, the Navigator resumes execution with the Judge’s feedback integrated into its context, enabling self-correction; if validation succeeds, the episode concludes and the answer is returned. 
At the \orchestrator level, the Validator’s assessment is combined with the Navigator’s final report, allowing the \orchestrator to decide whether to accept, refine, or replan the outcome.

\section{Evaluation Methodology}

We evaluate our system on four major benchmarks spanning web, desktop, and mobile environments: WebVoyager \cite{he_webvoyager_2024}, WebArena \cite{zhou_webarena_2024}, OSWorld \cite{xie_osworld_2024}, and AndroidWorld \cite{rawles_androidworld_2025}. All experiments employ models without any task-specific fine-tuning, isolating the contribution of our hierarchical agent architecture from model improvements. This experimental design demonstrates that superior performance can be achieved through careful system design and agent orchestration alone, independent of model scale or domain adaptation.

\subsection{Benchmarks}

Here, we outline the main characteristics of the benchmarks we use and briefly describe each of them. Comprehensive details, including task distributions, evaluation metrics, and corrections to prior evaluation inconsistencies, are presented in Appendix~\ref{app:benchmark_details}.

\paragraph{Key benchmark characteristics.}
\textbf{Multi-step reasoning}: Tasks require sequential actions with conditional branching based on intermediate observations, assessing the agent's ability to plan and adapt over long horizons. 
\textbf{Real-world environments}: Benchmarks rely on realistic environments, such as actual websites (WebVoyager, WebArena), production desktop applications (OSWorld), and authentic mobile apps (AndroidWorld) rather than simulators, exposing agents to dynamic content, varied layouts, and real-world edge cases. 
\textbf{High-precision visual grounding}: Success depends on accurate localization of UI elements within pixel-dense screenshots, where small coordinate errors cause action failures, demanding tight integration of vision and language reasoning. 
\textbf{Diverse interaction modalities}: The benchmark suite spans mouse/keyboard control (desktop), touch gestures (mobile), and hybrid web interactions, ensuring that our orchestration framework generalizes across fundamentally different action spaces and environment dynamics.

\paragraph{WebVoyager} consists of 643 tasks spanning 15 popular websites in its original formulation, including e-commerce, travel, and information platforms (e.g., Amazon, Booking.com, ArXiv). These tasks require complex agent interactions such as visually grounded information retrieval, comparison, and multi-step form completion.
However, the dynamic nature of live websites introduces instability that can render tasks obsolete. To ensure experimental comparability, we adopted the curated 590-task subset established by Magnitude \cite{magnitude_webvoyager}. Details on access restriction mitigations are available in Appendix \ref{app:wv-details}.

\label{benchmarks_webarena}
\paragraph{WebArena} is a suite of 812  tasks designed to test navigation capabilities across diverse web environments, including an e-commerce site, a social forum, a GitLab instance, a content management system, and a map interface. Refinements were made to the original WebArena implementation (see Appendix \ref{app:wa-details}).

\paragraph{OSWorld} spans 369 real Computer Use tasks on Ubuntu systems, spanning production applications such as LibreOffice, GIMP, Chrome, Thunderbird, VS Code, and VLC. Tasks test realistic workflows including document editing, image manipulation, email management, and multi-application coordination and are scored through deterministic programmatic checks.
We focus on the \textbf{Foundation E2E GUI} category, which constrains the agent’s action space to human-performable GUI operations: mouse clicks, drags, keyboard inputs, and shortcuts without calling APIs or executing code. This setting is the most representative of true Computer Use capability, as it requires agents to perceive and act directly on arbitrary interfaces rather than relying on handcrafted integrations.
By contrast, the broader OSWorld All category permits code-level operations (e.g., Python snippets or API calls) that can bypass interface-level reasoning, potentially inflating scores through tool-specific shortcuts rather than genuine GUI understanding.
We therefore evaluate \surfer in the stricter Foundation E2E GUI regime and compare it against competitors within that category, emphasizing generalization to unseen applications and fidelity to human interaction. 
For context, the current highest score in the “All” category is held by Agent S3 at 69.9\% accuracy \cite{gonzalezpumariega2025unreasonableeffectivenessscalingagents}.

\paragraph{AndroidWorld} evaluates mobile agent capabilities across 116 tasks spanning the Android OS itself and 20 real-world applications. These tasks require touch-based interactions, app navigation, and multi-app workflows verified through Android Debug Bridge (ADB)-based state inspection. Each task is scored with a verification metric in $\{0,0.5,1\}$, corresponding to failure, partial success, and success, respectively.
Importantly, the Android Emulator provides access to the accessibility tree (a11y), which agents can leverage to perform tasks. To better demonstrate the performance of our unified agent, we deliberately avoid using this accessibility tree and instead rely solely on visual (screenshot-based) inputs.

\subsection{Configuration}

\begin{table}[h!]
\caption{Model configuration across benchmarks.}
\centering
\vspace{0.3em}
\renewcommand{\arraystretch}{1.2}
\begin{tabular}{lcccc}
\toprule
\textbf{Benchmark} & \textbf{Orchestrator} & \multicolumn{3}{c}{\textbf{Navigator}} \\
\cmidrule(lr){3-5}
 & & \textbf{Policy} & \textbf{Judge} & \textbf{Localizer} \\
\midrule
WebVoyager & o3 & Claude Sonnet 4.5 & GPT 4.1 & \holo 7B \\
WebArena & o3 & Claude Sonnet 4.5 & o3 & \holo 72B \\
OSWorld & None & Claude Sonnet 4.5 & o3 & \holo 72B \\
AndroidWorld & None & o3 & o3 & \holo 72B \\
\bottomrule
\end{tabular}
\label{tab:model_config}
\end{table}

For each benchmark, we adapt the model configuration and component selection to the specific environment and task complexity, as determined through ablation studies (see Table \ref{tab:model_config}. 
In WebVoyager and WebArena, the \orchestrator operates for up to 20 steps, while the Navigator may take up to 50 steps to complete navigation subtasks.
For OSWorld and AndroidWorld, the agent runs without an \orchestrator, relying solely on the Navigator for both planning and execution. 
In OSWorld, we enforce a minimum of 15 and a maximum of 100 steps to prevent premature termination. Once the upper limit is reached, the agent’s memory is cleared without resetting the environment enabling continued progress while keeping the context length bounded. 
In AndroidWorld, the step limit is set to 150, reflecting the higher difficulty of certain tasks (e.g., \texttt{OsmAndTrack} which has an empirically determined optimal horizon of roughly 60 steps).

\section{Results}

In this section, we describe both qualitative and quantitative main results obtained on the four benchmarks.

\subsection{WebVoyager}
\label{sec:webvoyager}

\begin{figure}[h!]
    \centering
    \includegraphics[width=\linewidth, trim=0 5cm 0 5cm, clip]{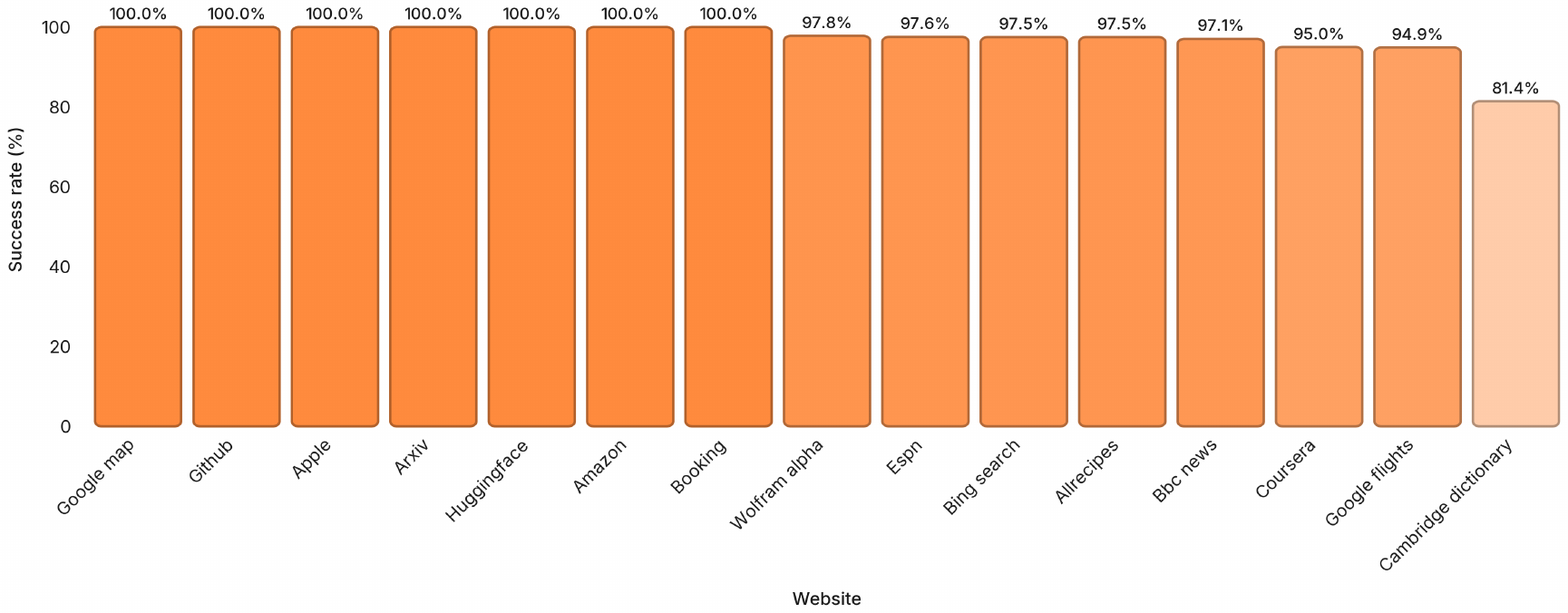}
    \caption{Per-website performance of \surfer on the WebVoyager benchmark.}
    \label{fig:wv-sr-by-domain}
\end{figure}

\surfer establishes a new state of the art on the WebVoyager benchmark, achieving a 97.1\% success rate and surpassing the previous best performance of~93.9\%~\cite{magnitude_webvoyager}. This strong performance is consistent across nearly all tested websites (see Figure \ref{fig:wv-sr-by-domain}), with the exception of the Cambridge Dictionary domain, where anti-bot measures such as CAPTCHAs hindered execution. \surfer achieves a perfect 100\% pass@10, effectively saturating the benchmark using test-time scaling. In a localizer ablation study, substituting \holo 7B with UI-TARS 7B \cite{qin_ui-tars_2025} reduced performance to 94.7\%, confirming that \surfer’s gains derive from the combination of high-quality components and effective orchestration.

\subsection{WebArena}

\begin{figure}[h!]
    \centering
    \includegraphics[width=\linewidth]{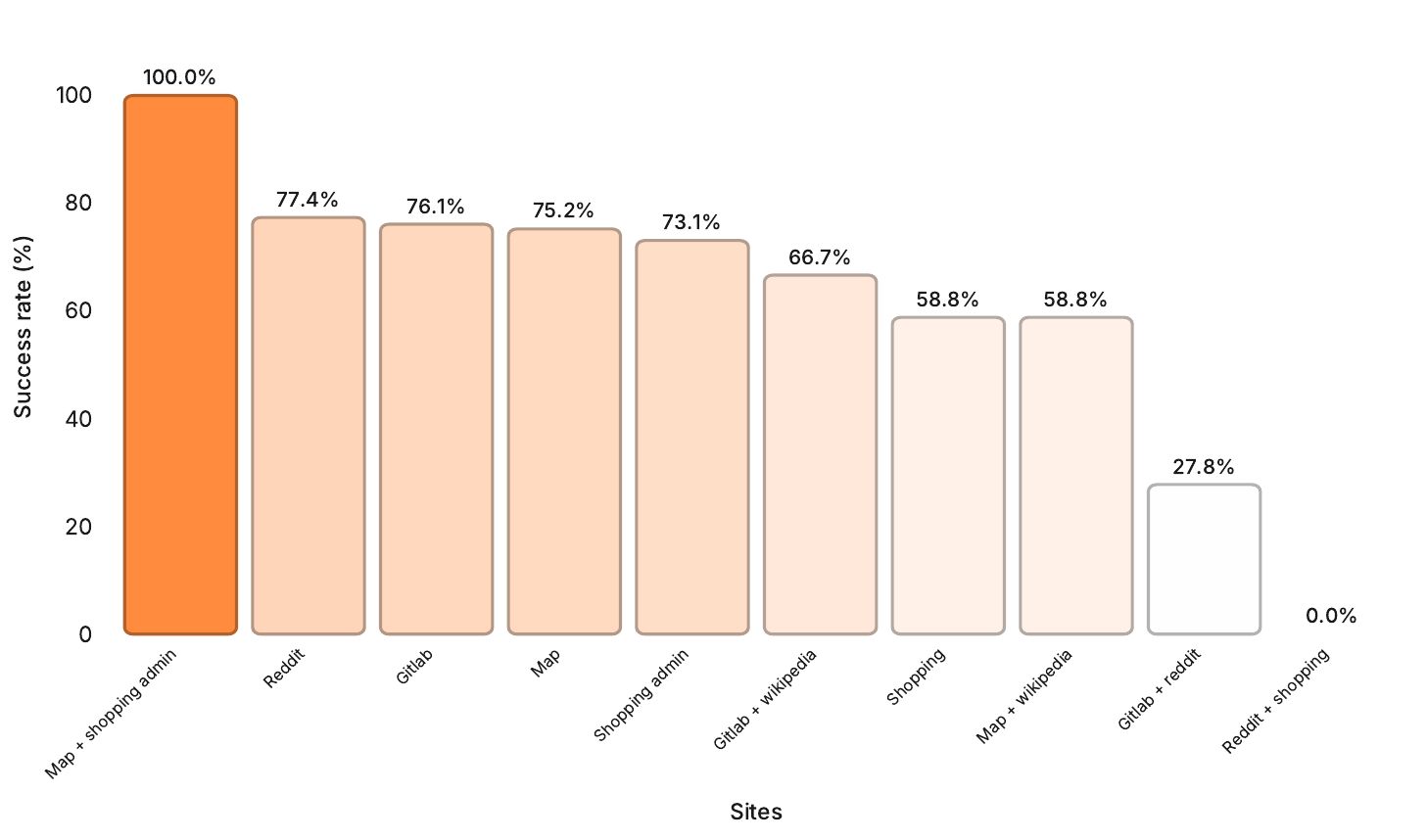}
    \caption{Per-domain performance of \surfer on the WebArena benchmark.}
    \label{fig:wa-sr-by-domain}
\end{figure}

\surfer reaches a new state-of-the-art with a $\text{pass}@1$ success rate of $69.6\%$ on WebArena. The agent performed robustly on social media tasks, e.g., $77\%$ on Reddit, while struggling with e-commerce workflows, averaging only $58\%$ on shopping sites. Many tasks in this domain remain challenging, see Figure \ref{fig:wa-sr-by-domain}.

\paragraph{Test-time scaling.} Sampling multiple independent trajectories of \surfer leads to a substantial performance gain from 69.6\% with $\text{pass}@1$ to 84.9\% with $\text{pass}@10$. Scaling parallel trajectories primarily expands task coverage, revealing what the agent \textit{can} do rather than increasing single-run reliability. The diversity of successful paths indicates that most residual failures arise from local exploration traps rather than systematic reasoning errors. High task coverage offers a valuable signal about the agent’s effective action space and highlights its potential for reinforcing successful behavioral patterns through model training.

\subsection{OSWorld}

\begin{figure}[h!]
    \centering
    \includegraphics[width=\linewidth]{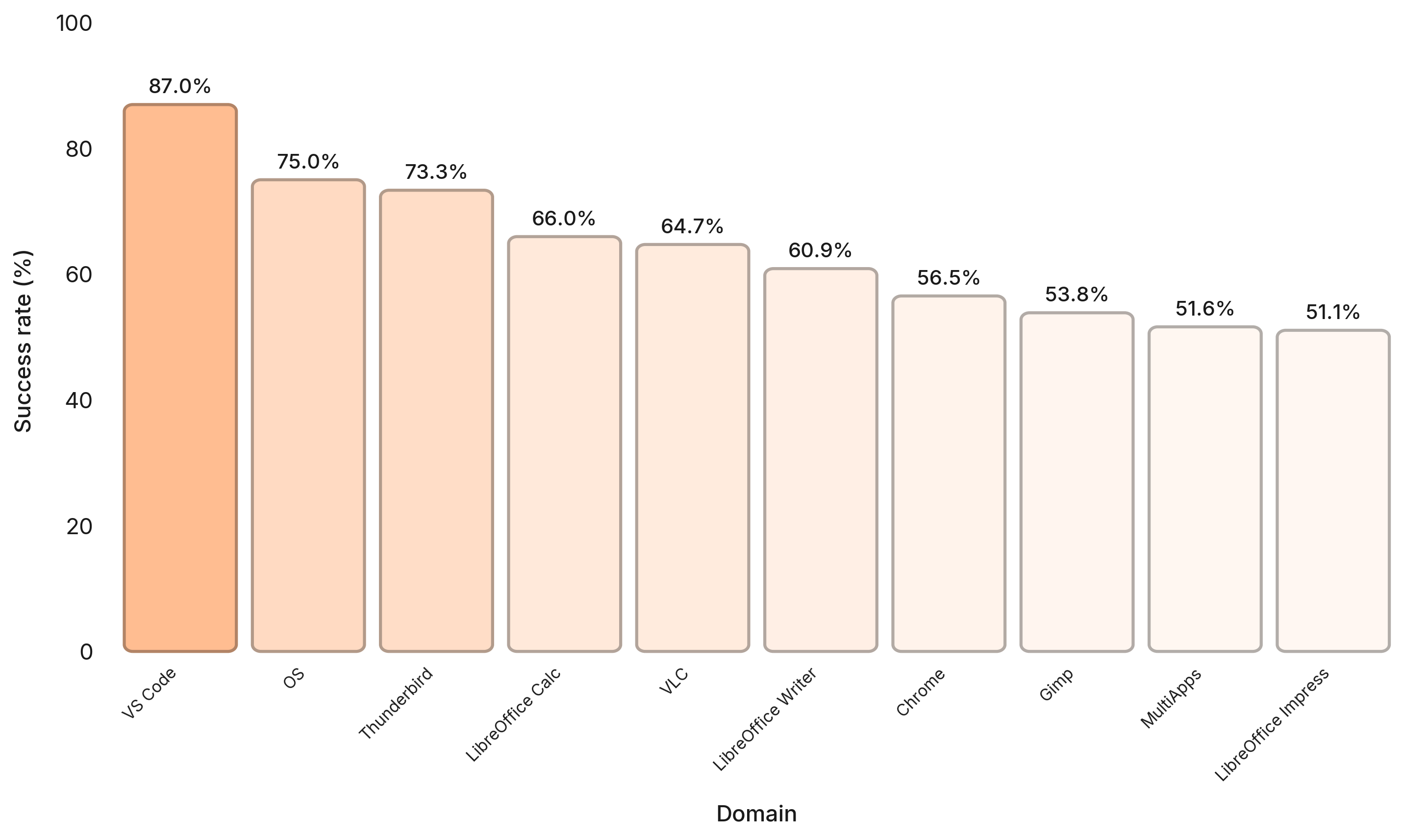}
    \caption{\surfer performance per task category on the OSWorld benchmark.}
    \label{fig:osw_per_domain}
\end{figure}

\surfer achieves a state-of-the-art success rate of 60.1\% on OSWorld in the Foundation E2E GUI category. 
With five attempts ($\text{pass}@5$), performance rises to 72.0\%, closely matching the human baseline of 72.4\%. At ten attempts ($\text{pass}@10$), \surfer surpasses human performance with 77.0\%. 
In Figure \ref{fig:osw_per_domain} we report success rates across task categories; in all of them, \surfer exceeds the accuracy of 50\%, performing especially well on programming-related tasks in environments like VSCode and OS. Interestingly, \surfer completed several tasks labeled as infeasible by human evaluators, which were excluded from our success rate (see Figure~\ref{fig:korean-surfer}).

\paragraph{Localizer ablation.} Within the same agentic framework, \holo 72B Localizer achieved the highest performance, reaching 60.1\% compared to 58.4\% for \holo 7B and 56.9\% for UI-TARS 7B \cite{qin_ui-tars_2025}. This confirms the importance of accurate spatial grounding for GUI reasoning: \holo generalizes more effectively to diverse Computer Use interfaces, enabling precise GUI interaction.

\subsection{AndroidWorld}

\surfer achieves an accuracy of $87.1\%$ across the 116 tasks, surpassing all previous approaches relying solely on visual interaction \cite{gbox_android_world,k2_agent_2025}. 
\begin{figure}[h!]
    \centering
    \includegraphics[width=0.35\linewidth]{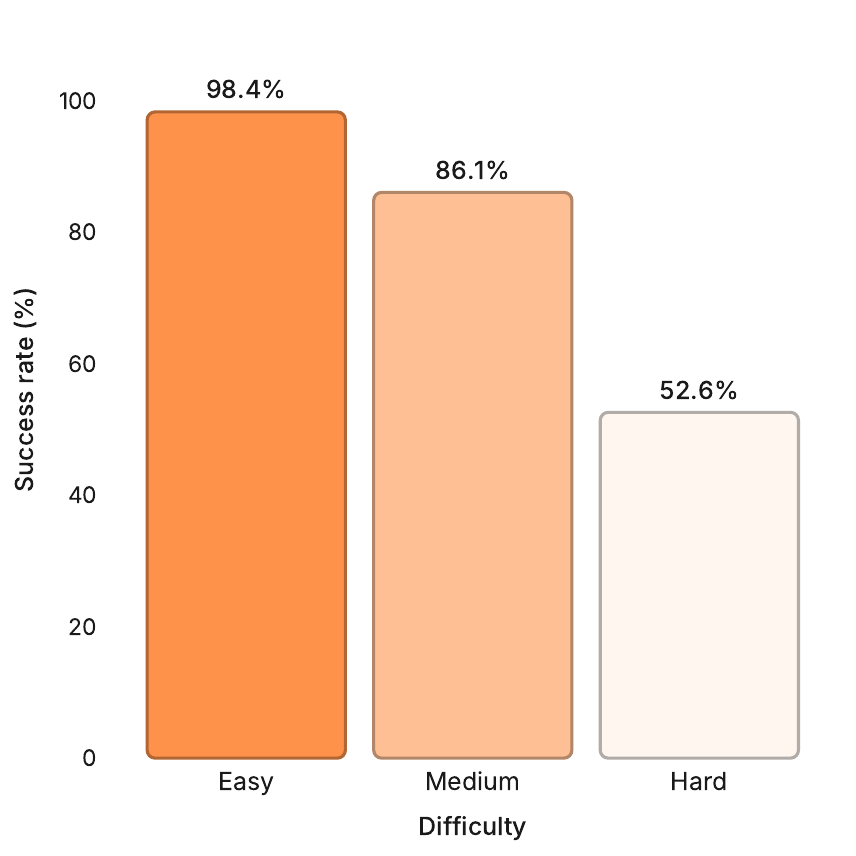}
    \caption{\surfer accuracy per difficulty category on AndroidWorld.}
    \label{fig:androidworld2_difficulty_accuracy}
\end{figure}
Figure~\ref{fig:androidworld2_difficulty_accuracy} shows the agent’s success rate across difficulty levels as defined in the original AndroidWorld paper. It achieves a near-perfect performance on Easy tasks (98.4\%), maintains strong results on Medium tasks (86.1\%), and shows a notable drop on Hard tasks (52.6\%). This trend indicates that while \surfer generalizes well to moderately complex interactions, performance decreases on tasks requiring long-horizon reasoning or intricate multi-step coordination.

\begin{figure}[h!]
    \centering
    \includegraphics[width=\linewidth]{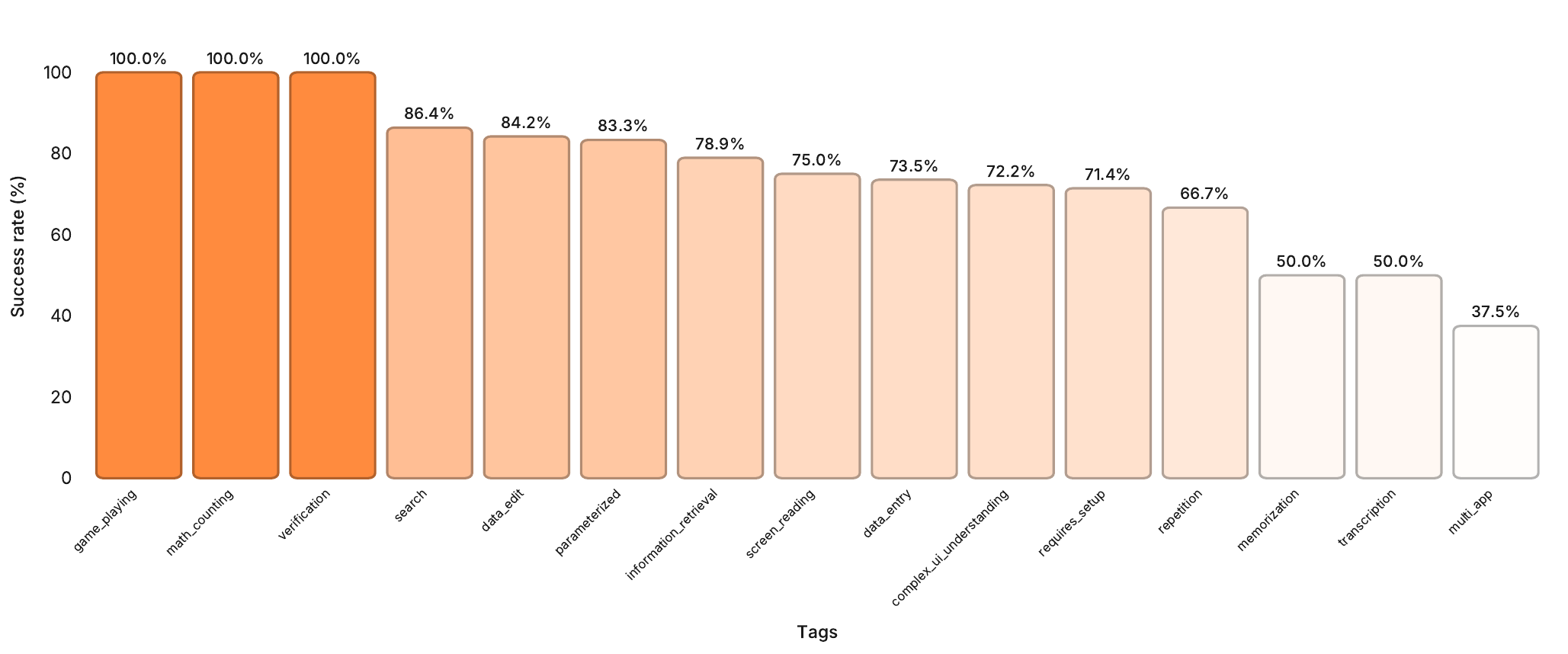}
    \caption{\surfer accuracy by task category on AndroidWorld.}
    \label{fig:androidworld2_tags_accuracy}
\end{figure}

\paragraph{Performance per category.}
Performance varies across categories (Figure~\ref{fig:androidworld2_tags_accuracy}), with memorization and transcription (both $50\%$) identified as the most difficult due to limited memory and text handling, and multi-app tasks ($37.5\%$) remaining the primary challenge for robust cross-application reasoning.

\paragraph{Test-time scaling.}
Allowing a small retry budget yields consistent improvements on AndroidWorld: $\text{pass}@1$ reaches $87.1\%$, $\text{pass}@2$ improves to $90.5\%$ ($+3.4\%$), and $\text{pass}@3$ climbs to $93.1\%$ ($+6.0\%$ over $\text{pass}@1$). These gains indicate that a modest number of parallel attempts recovers many near-miss failures, suggesting that most errors arise from stochastic perception or planning rather than fundamental limitations.

\paragraph{Localizer ablation.}
Substituting \holo with UI-TARS reduces pass@1 success from $87.1\%$ to $81.9\%$ ($-5.2\%$) on AndroidWorld, indicating the localizer as performance bottleneck. A drop in accuracy in localization of interactions with small targets and iconographic widgets cascades into incorrect actions. This ablation underscores that the spatial grounding from \holo is crucial for end-to-end reliability.

\paragraph{Implications.}
\surfer effectively manages complex multi-app, multi-step workflows. The remaining challenges lie in (i) real-time perception, (ii) maintaining memory over long sequences, and (iii) stronger GUI semantics—particularly learning app-specific visual conventions and icon mappings.

\section{Key Insights and Path Forward}
Our empirical evaluation reveals critical factors for agent success. Prompt engineering proves surprisingly impactful, with minor wording changes yielding 5--10\% accuracy swings, underscoring the sensitivity of LLM reasoning to input formatting. Model variance remains substantial even at temperature 0 for the judge, necessitating multi-sampling strategies to achieve robust performance. Persistent context across subtasks reduces navigation steps required by 30--40\%, proving essential for multi-goal tasks that build on prior subtasks. Multi-stage validation intercepts 15--20\% of errors before propagation, while \orchestrator's hierarchical decomposition provides natural retry boundaries and improved interpretability.

Despite achieving human-level performance on complex tasks, several bottlenecks constrain practical deployment. Stochastic model outputs require expensive multi-sampling for reliability, with \orchestrator costs reaching~\$1--5 per complex task when using frontier reasoning models. Even state-of-the-art localizers fail on~5--8\% of UI elements due to dynamic content and ambiguous descriptions. Long-horizon tasks exceeding~50 steps face context window limits and compounding errors, while LLM-based evaluation itself carries~5--10\% error rates that complicate benchmarking.

Our results demonstrate that proper agent orchestration with fixed, general-purpose models can achieve state-of-the-art accuracy, with modular designs generalizing across web, desktop, and mobile environments.

While the principles of agent orchestration are approaching maturity, the remaining bottlenecks -- cost, variance, and speed -- limit practical, real-world deployment. 
The path forward lies in resolving these challenges. To address this, we are focusing on developing a new family of smaller, specialized models, to achieve comparable, if not superior, performance at a fraction of the current cost, reaching Pareto optimality~\cite{andreux_surfer-h_2025}.

\clearpage

\printbibliography

\newpage
\appendix

\section{Appendix}
\label{app:benchmark_details}

\subsection{WebVoyager Evaluation}
\label{app:wv-details}

\begin{itemize}
    \item \textbf{Description}: 590 tasks across 15 websites: Amazon, Apple, Google Flights, Booking.com, ArXiv, GitHub, Hugging Face, Coursera, BBC News, Cambridge Dictionary, Allrecipes, Google Maps, Bing Search, ESPN and Wolfram Alpha.
    \item \textbf{Task types}: Information retrieval, comparison and navigation.
    \item \textbf{Evaluation}: We updated the evaluation protocol by replacing the single GPT-4V judge with a majority vote over 3 GPT-4o judge calls (temperature 0, using last 5 screenshots), retaining the original WebVoyager evaluation prompt. This ensemble method is designed to reduce variance, mitigate single-judge bias, and prevent spurious results from isolated errors, thereby yielding more reliable and robust evaluations.
    \item \textbf{Environment}: Live websites on Selenium-controlled Chrome. Running this benchmark introduces practical challenges, such as bot-detection mechanisms like CAPTCHAs and IP-based access blockers. To mitigate these restrictions, we employed proxy rotation and redirected all Google-dependent tasks to Bing Search, as in~\cite{andreux_surfer-h_2025}, to ensure uninterrupted execution.
\end{itemize}

\subsection{WebArena Evaluation}
\label{app:wa-details}

\label{webarena_details}
\begin{itemize}
\item \textbf{Description}: 812 tasks across self-hosted 6 websites: GitLab, Reddit, an E-commerce website called OneStopShop, an online store content management system (CMS), OpenStreetMap and the English Wikipedia.
\item \textbf{Task types}: Information retrieval, comparison, navigation, multi-step  and multi-websites actions, involving form filling and search across one or two websites. To improve grounding and consistency, we added lightweight, category-specific initialization prompts, loosely inspired by OpenAI’s per-site prompting approach~\cite{openai_cua_eval_2025}.
\item \textbf{Evaluation}: modular evaluation framework consisting of three criteria that can be combined or used independently:
\begin{itemize}
\item URL Match: The agent's final URL must match a predefined target.
\item HTML Artifact: A required success artifact must be present in the final page's DOM.
\item Model-based Assessment: A language model evaluates the correctness of the final output using majority voting with 5 GPT 4.1 judges.\\
\end{itemize}
\item \textbf{Environment}: Controlled web servers with fixed initial states accessed using Selenium-controlled Chrome.
\end{itemize}

\subsubsection{Methodological Difficulties with WebArena.}

\paragraph{Technical challenges.}
While WebArena is a valuable benchmark, its evaluation poses several methodological challenges. The benchmark requires self-hosting of its websites, a process that is not seamless and which required manual intervention, notably to resolve issues in the OpenStreetMap environment. Its highly stateful design further complicates reproducibility, as tasks are interdependent and their outcomes can be affected by residual states from previous runs. These characteristics also make large-scale parallel execution impractical, substantially increasing the computational cost and time required to obtain robust metrics

\paragraph{LLM-as-a-Judge.}
The benchmark's original evaluation framework integrated programmatic checks of the final state with a single LLM-based assessment. We improved its rigor by replacing heuristic string matching (e.g., exact-match or keyword checks) with an ensemble of five independent GPT-4.1 judges, aggregated by majority vote. This approach reduces variance, mitigates single-judge bias, penalizes false positives in $\textit{must include}$ tasks where substring matching previously sufficed, and allows minor, semantically irrelevant variations in $\textit{exact match}$ tasks that would otherwise fail under literal comparison. Given that WebArena contains 176 $\textit{must include}$ tasks versus only 45 $\textit{exact match}$ tasks, our methodology provides a more balanced and reliable assessment.

\paragraph{Task Corrections.}
We conducted a manual review of the benchmark dataset \cite{webarena_github}, resulting in the correction of 71 tasks with erroneous labels. The scope of these modifications ranged from minor typographical fixes to the resolution of critical logical inconsistencies that fundamentally affected evaluation outcomes. A summary of these corrections is provided in Table \ref{tab:fuzzy_corrections_summary}. 
For a transparent comparison, we conducted an ablation study without these manual fixes, relying solely on a Large Language Model (LLM) as the judge for string comparison. In this configuration, our approach achieves a success rate of $67.4\%$, a result that still surpasses the previous state-of-the-art performance.

\begin{table}[H]
\centering
\footnotesize 
\begin{threeparttable}
\caption{Summary of WebArena corrections with examples.}
\label{tab:fuzzy_corrections_summary}
\renewcommand{\arraystretch}{1.15} 
\setlength{\tabcolsep}{4pt}       
\begin{tabular}{
  >{\raggedright\arraybackslash}p{0.21\textwidth} 
  >{\raggedright\arraybackslash}p{0.52\textwidth} 
  >{\raggedright\arraybackslash}p{0.25\textwidth}
}
\toprule
\textbf{Category} & \textbf{Description} & \textbf{Example correction} \\
\midrule

Data accuracy corrections & 
Rectifying errors in numerical values (e.g., quantities, measurements, prices), and making corrections to specific names or entities to reflect the correct data. & 
Corrected order count from ``24 orders'' to ``\textbf{21 orders}'' (Task~50). \\

Typographical and spelling fixes & 
Addressing simple spelling errors in task fields such as \texttt{intent}, \texttt{intent\_template}, or \texttt{instantiation\_dict} to improve clarity. & 
Corrected ``canlled'' to ``\textbf{cancelled}'' (Task~202). \\

URL and task focus updates & 
Updating URLs, parameters, and location references to ensure that tasks point to the intended content or have a more precise focus. & 
Updated URL parameter from \texttt{sort=created\_asc} to \texttt{sort=created\_date} (Task~45). \\

Evaluation flexibility improvements & 
Replacing specific data points (e.g., domains) with flexible placeholders to make the evaluation process more robust against variations in expected outputs. & 
Replaced a specific domain with the placeholder: 

\texttt{\textless fuzzy\_general\_domain\textgreater} (Task~293). \\

Consistency and formatting edits & 
Standardizing date formats, ensuring consistent phrasing, and adding URL parameters to control the number of items displayed for a uniform evaluation environment. & 
Standardized proximity phrasing by replacing ``around'' with ``\textbf{near}'' (Task~378). \\

\bottomrule
\end{tabular}
\end{threeparttable}
\end{table}

\subsection{OSWorld Evaluation}
\label{app:osw-details}

\begin{itemize}
    \item \textbf{Description}: 369 tasks on Ubuntu Desktop. They involve different applications: LibreOffice Calc, LibreOffice Impress, LibreOffice Writer, Thunderbird, Gimp, OS, Chrome, VS Code, VLC. The task can involve several applications at the same time.
    \item \textbf{Task types}: document editing, image manipulation, email, file management, settings modifications, data analysis, coding, etc. The complexity of the tasks varies significantly (10-100 steps).
    \item \textbf{Evaluation}: Programmatic checks of final file/application state.
    \item \textbf{Environment}: AWS-hosted Ubuntu VMs, 1920x1080, VNC control.
\end{itemize}

\subsection{AndroidWorld Evaluation}
\label{app:aw-details}

\begin{itemize}
    \item \textbf{Description}: 116 tasks on an Android Emulator (Pixel 6 device). Tasks are specified as parameterized templates instantiated at runtime, requiring the agent to interact with apps such as Markor (notes), VLC (video), OpenTracks (activity tracker), Simple Calendar, Tasks.org, SMS, Contacts, Files, Camera, Audio Recorder, OsmAnd (maps), Retro Music Player, Recipe and Expense managers, Clock, and system settings, emphasizing multi‑app, multi‑steps workflows.
    \item \textbf{Task types}: Data entry/edit, information retrieval and search/navigation, screen reading and transcription, math/counting, verification, memorization/repetition, multi-app workflows, complex UI understanding, parameterized inputs, requires setup, game playing.
    \item \textbf{Evaluation}: Programmatic checks of final device and application state via the Android Debug Bridge. For information retrieval tasks, validation through exact or fuzzy matching. Each task is assigned a score of $1$, $0.5$, or $0$ for success, partial success, or failure, respectively.
    \item \textbf{Environment}: Dockerized Android emulator (Pixel 6 device, 1080×2400 pixels, API level 33). Due to SIM card constraints, SMS-related tasks were executed outside the Docker containers.
    
\end{itemize}

\section{Examples}
\label{app:benchmark-examples}
\subsection{OSWorld Example}
\label{app:osworld-example}

\begin{figure}[h!]
    \centering
    \includegraphics[width=0.82\linewidth]{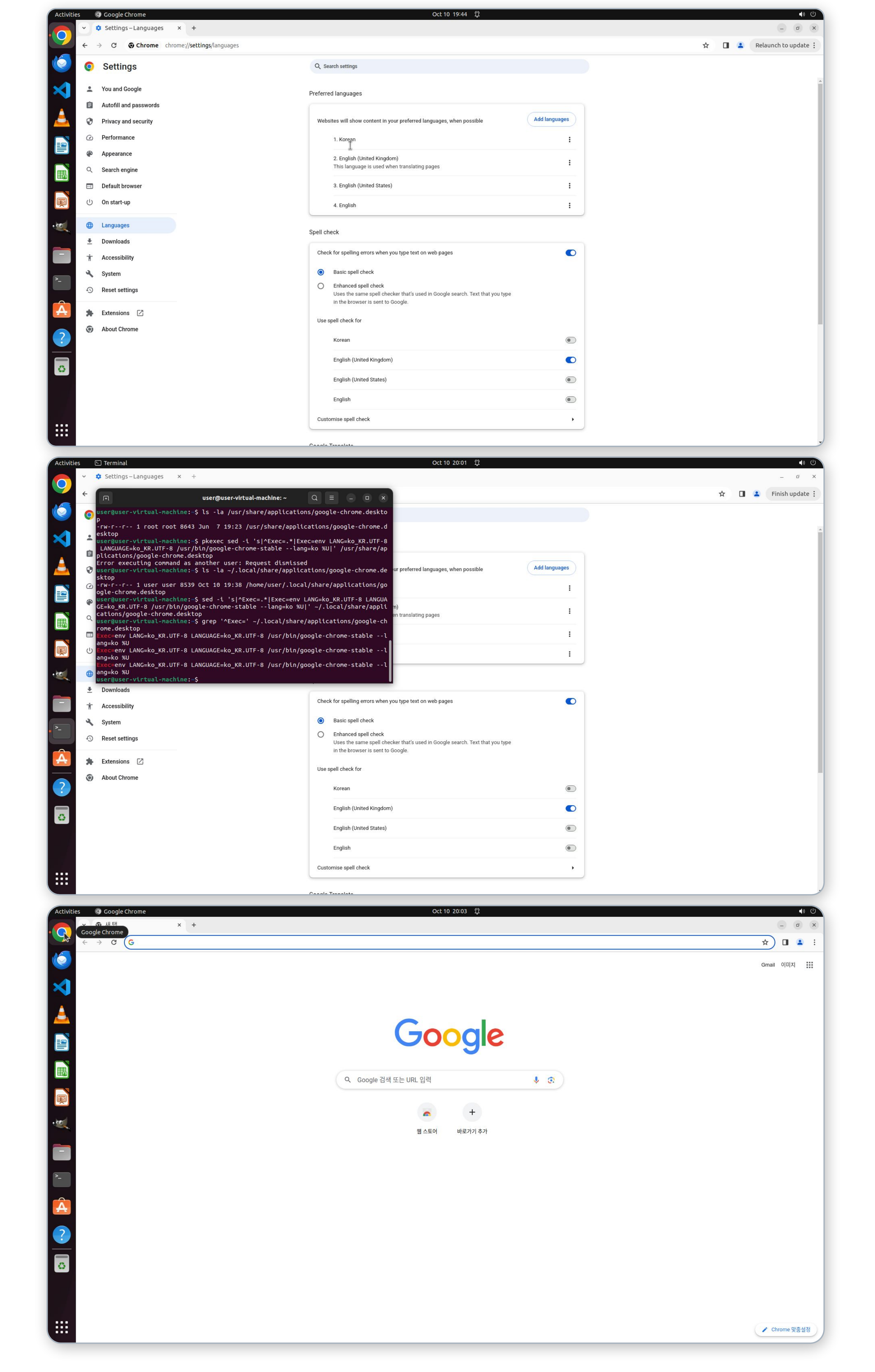}
    \caption{\surfer switching Chrome’s language to Korean via the terminal.}
    \label{fig:korean-surfer}
\end{figure}

To illustrate \surfer’s adaptive behavior in desktop environments, we highlight one notable case where the agent successfully solved a task labeled as infeasible by human evaluators. For instance, it successfully changed Chrome’s interface language to Korean--a task deemed impossible due to the absence of a visible language selector (see Figure \ref{fig:korean-surfer}). Instead of reporting failure, \surfer opened a terminal, executed system commands via visual interaction, and relaunched Chrome, showcasing adaptive reasoning beyond predefined task boundaries through the integration of interface understanding and system-level knowledge.

\subsection{AndroidWorld Examples}

\begin{figure}[h!]
  \centering
  \begin{subfigure}{0.485\linewidth}
    \centering
    \includegraphics[width=0.9\linewidth]{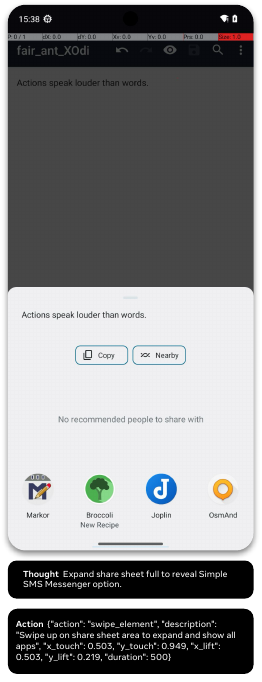}
    \caption{\textbf{(Success) \textsc{MarkorCreateNoteAndSms}}: The agent creates a new note in Markor with the text \textit{``Actions speak louder than words.''}, then shares it via Simple SMS Messenger. 
    This requires cross-app coordination (Markor → SMS), which \surfer executes by creating the file, using the share action, and selecting the SMS app.}
    \label{fig:markor-success}
  \end{subfigure}\hfill
  \begin{subfigure}{0.485\linewidth}
    \centering
    \includegraphics[width=0.9\linewidth]{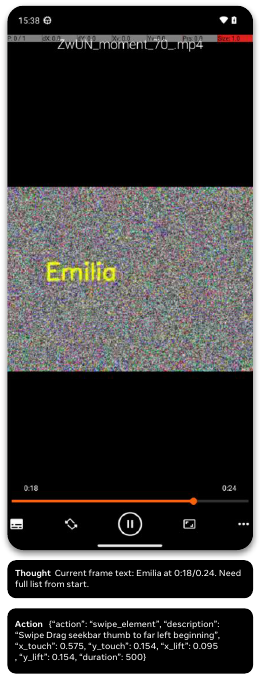}
    \caption{\textbf{(Failure) \textsc{MarkorTranscribeVideo}}: The agent transcribes per-frame text from a VLC video into a Markor note. 
Despite reaching the correct context, limited temporal memory and sparse frame capture result in partial transcription and omissions, 
revealing difficulties in maintaining continuity over extended visual sequences.}
    \label{fig:markor-failure}
  \end{subfigure}
\caption{Representative success and failure cases on AndroidWorld. \surfer demonstrates robust cross-app reasoning but still struggles with long-horizon temporal perception.}
  \label{fig:androidworld-examples}
\end{figure}

To contextualize aggregate metrics, we highlight one complex success (Figure \ref{fig:markor-success}) and one representative failure (Figure \ref{fig:markor-failure}).

\end{document}